\documentclass[letterpaper, 10 pt, conference]{ieeeconf}  

\IEEEoverridecommandlockouts                              

\usepackage{bm}
\usepackage{graphicx} 
\usepackage[linesnumbered,lined,algoruled]{algorithm2e}
\usepackage{booktabs}
\usepackage{multirow}
\usepackage{cite}
\usepackage{amsmath} 
\usepackage{amssymb}  

\title{\LARGE \bf
Vision-based Distributed Multi-UAV Collision Avoidance via Deep Reinforcement Learning for Navigation
}

\author{Huaxing Huang$^{1}$, Guijie Zhu$^{1}$, Zhun Fan$^{1,*}$, Hao Zhai$^{1}$, Yuwei Cai$^{1}$, Ze Shi$^{1}$, Zhaohui Dong$^{1}$, Zhifeng Hao$^{1}$
\thanks{*This work is supported by the Key Lab of Digital Signal and Image Processing of Guangdong Province, and by the National Defense Technology Innovation Special Zone Project  (193-A14-226-01-01).}
\thanks{$^{1}$ Key Lab of Digital Signal and Image Processing of Guangdong Province, College of Engineering, Shantou University.
{\tt\small zfan@stu.edu.cn}}%
}

\begin{document}

\maketitle
\thispagestyle{empty}
\pagestyle{empty}

\begin{abstract}

    Online path planning for multiple unmanned aerial vehicle (multi-UAV) systems is considered a challenging task. It needs to ensure collision-free path planning in real-time, especially when the multi-UAV systems can become very crowded on certain occasions. In this paper, we presented a vision-based decentralized collision-avoidance policy for multi-UAV systems, which takes depth images and inertial measurements as sensory inputs and outputs UAV's steering commands. The policy is trained together with the latent representation of depth images using a policy gradient-based reinforcement learning algorithm and autoencoder in the multi-UAV three-dimensional workspaces. Each UAV follows the same trained policy and acts independently to reach the goal without colliding or communicating with other UAVs. We validate our policy in various simulated scenarios. The experimental results show that our learned policy can guarantee fully autonomous collision-free navigation for multi-UAV in the three-dimensional workspaces with good robustness and scalability.

\end{abstract}

\section{INTRODUCTION}

In recent years, multi-robot navigation has received much interest in, e.g., robotics and artificial intelligence. It has a wide range of applications in robotic search and rescue [1], agricultural irrigation [2] and delivery [3], etc. The general purpose of robot navigation is to identify an optimal path from a starting point to a target point in an environment while avoiding obstacles. There have been many works that have proposed efficient methods for multiple Unmanned Ground Vehicles (multi-UGV) path planning [4], [5], [6]. However, the path planning algorithms for UAVs differ from those of UGVs in that the collision avoidance problem needs to be solved in the three-dimensional workspaces, which poses much more difficulties and challenges.

Some researchers propose solutions to multi-UAV collision avoidance systems based on centralized methods, which assume that the actions of all UAVs are determined by a central server that can access global information about UAVs (e.g., all UAVs' locations) and workspaces (e.g., a grid map) [7], [8]. However, the centralized multi-UAV systems rely heavily on communication between UAVs and the central server. In addition, they require a high computational cost for scheduling, which makes them difficult to be deployed in practice.

Compared with centralized methods, some existing works proposed decentralized methods for multi-UAV collision avoidance systems, where each UAV plans the optimal path only through its own sensors and limited onboard computing resources. However, many of these methods require mutual communication between UAVs to share information (e.g., each UAV's GPS location [9] or planning trajectory [10]), making it difficult to maintain the system's robustness in a limited communication environment. There are also methods that leverage the information of observable objects for collision avoidance decision-making [11]. However, these methods assume that each UAV can obtain the accurate positions of surrounding objects through its own sensors. Still, this assumption does not hold in the real world owing to imperfect sensing. Wang et al. [12] propose a learning-based decentralized collision avoidance method without the perfect sensing assumption, but it often requires identifying objects first and then roughly estimating distances before making collision avoidance decisions when deployed in practice, which hurts the real-time performance of the system. 

\begin{figure*}                                                            
    \centering
    \includegraphics[width=505pt]{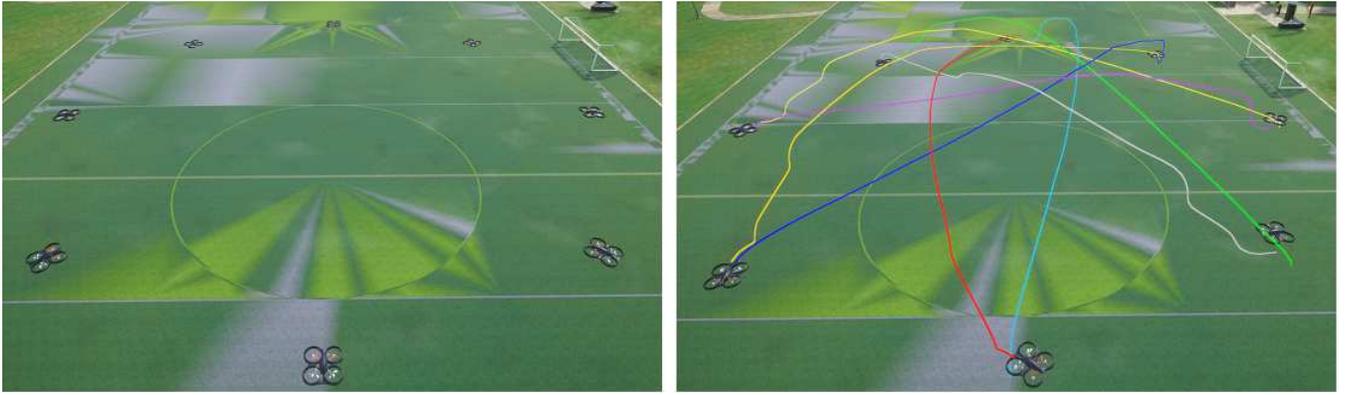}
    \caption{\textbf{left:} The Airsim simulator used in this paper.  \textbf{right:} UAVs trajectories in the Circle scenario using our learned policy.}
    \label{fig1}
\end{figure*}

In this paper, we propose a vision-based decentralized collision-avoidance policy by deep reinforcement learning, which mainly leverages a vision sensor and
inertial measurements on-board to accomplish robust and scalable multi-UAV path planning without communicating with other UAVs. In addition, it does not require external positioning and computation or pre-built maps. Since the existing reinforcement learning methods already have low sample complexity, it is more inefficient to use pixel-based input directly. We use an autoencoder to train latent state representations and policy jointly. We demonstrate that the collision avoidance policy learned in the proposed method behaves robust and stable compared to the RL model that directly uses convolutional neural networks. The experimental results show that the learned policy can generate collision-free paths for multi-UAV systems and maintain its navigation performance when the density of UAVs in the environment increases. To the best of our knowledge, this is the first work to demonstrate a vision-based collision-avoidance policy learned for a multi-UAV system without communication.

\section{RELATED WORK}
Multi-UAV navigation has always been challenging in unknown, complex environments, especially without any external positioning and communication. We mention some closely related works here in this section. 

\textbf{Learning-based Navigation:}
With the powerful representation capabilities of deep-learning technology, some recent works propose to learn navigation policies directly from data. These policies are trained by imitating experts (e.g., human experts [13] or sampling-based motion planning algorithms [14]) from experiences collected in simulation [15] or directly in the real world [16]. One significant advantage of these approaches is that they can jointly optimize all model parameters, thus reducing the effort of tuning each component. However, it is challenging to hand-design a proper loss function for training the policy, and the expert trajectories are not guaranteed to be optimal in the training scenarios, which makes it difficult to converge to a robust and general solution. There are some other methods based on deep reinforcement learning (DRL). Faust et al. [17] use global information provided by the traditional planners to assist robots in learning collision-avoidance policy. Leiva et al. [18] train map-less navigation policies using two-dimensional point clouds generated by range sensors. Ma et al. [19] present a vision-based UAV collision-avoidance framework, which generates saliency maps through supervised learning, and then trains control policy through reinforcement learning to avoid obstacles in the three-dimensional workspaces. These methods describe the navigation problem as a markov decision process (MDP) and use sensor measurements to learn navigation policies directly. The DRL-based method can achieve mapless navigation, and has strong learning ability and low dependence on sensor accuracy.

\textbf{Decentralized Multi-Robot Collision Avoidance:}
In recent years, most of the works for decentralized multi-robot navigation are based on optimal reciprocal collision avoidance (ORCA) framework [11], which assumes that each robot can perfectly sense the shape, position, velocity of other robots and obstacles in order to avoid collisions [20], [21]. To alleviate the requirement of perfect sensing, Godoy et al. [22] take the means of communication to share the robots' velocity information. Several other limitations still exist in this method. For example, it often requires a fine parameter tuning for better performance, and introducing the communication network damages the robustness of the multi-robot system, especially in environments when communication is hindered. There has been increasing research on DRL-based multi-robot collision-avoidance systems [23], [24]. Each robot is deployed with multiple sensors to estimate the states of surrounding agents and obstacles for path planning but needs a complicated pipeline for states estimation. The closest works to this paper are [25] and its extended work [26], which optimize a fully decentralized collision-avoidance policy in two-dimensional workspaces with the raw sensor measurements, including two-dimensional laser scans and inertial measurements. However, they fail to apply in multi-UAV systems because UAV path planning has a higher DOF than UGV planning. Bastien et al. [27] have proved that a vision-based model is sufficient to generate organized swarm behavior without spatial representation and collision. Inspired by the above works, we demonstrate multi-UAV collision avoidance in three-dimensional workspaces with deep reinforcement learning from a vision-based input.

\begin{figure*}
    \centering
    \includegraphics[width=505pt]{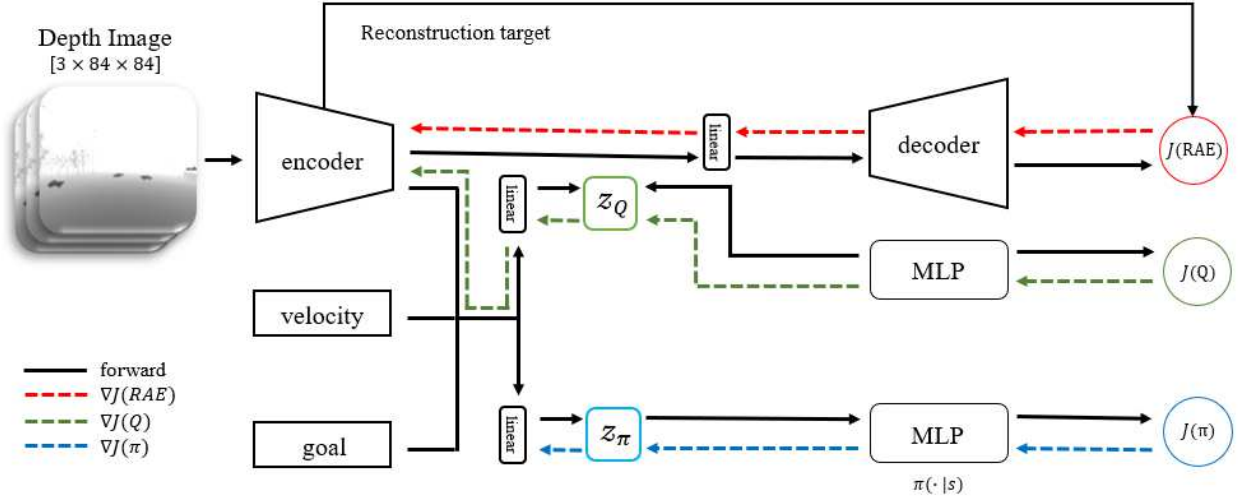}
    \caption{The architecture of the collision avoidance neural network. The network has the input of the depth images, current velocity, and relative goal position. Our method uses a regularized autoencoder to achieve dimensionality reduction on images and policy learning. The dotted lines denote the backward gradient propagations from different objective functions ($J(RAE)$, $J(Q)$ and $J(\pi)$). Note that we prevent the actor’s gradients from updating the convolutional encoder to achieve stable training from images.}
    \label{fig2}
\end{figure*}

\section{APPROACH}

In this section, we present the proposed method in detail. We introduce our deep reinforcement learning framework for multi-UAV systems first, then describe the neural network structure of the control policy. 

\subsection{Problem Formulation}
In our work, each UAV can only avoid collisions through its sensors, meaning the environment is only partially observable. Therefore, this kind of multi-UAVs collision avoidance problem is a partially observable markov decision process (POMDP).
The POMDP can be described as a 6-tuple $(\mathcal{S}, \mathcal{A}, \mathcal{T}, \mathcal{R}, \Omega, \mathcal{O})$, where $\mathcal{S}$ is the state set of the environment ($s \in \mathcal{S}$), all of which are partially observable, $\mathcal{A}$ is a set of actions($a \in \mathcal{A}$), $\mathcal{T}$ is the state transition function, which describes the probability of transition between states, $\mathcal{R}$ is the reward function$(\mathcal{S} \times \mathcal{A} \rightarrow \mathbb{R})$, $\Omega$ is a set of observations $(o \in \Omega)$ and $\mathcal{O}$ is the observation distribution given the state $(o \thicksim \mathcal{O}(s))$. 

In our formulation, we first assume a set of $N$ UAVs sharing an environment. In each timestep $t$, the $i$th UAV$(1 \leqslant i \leqslant N)$ receives an observation $\bm{o}_{i}^{t}$. It then executes an action $\bm{a}_{i}^{t}$ through calculation so that all UAVs are guaranteed collision-free move from the current position $\bm{p}_{i}^{t}$ towards the goal position $\bm{g}_{i}^{t}$. In the whole navigation process, each UAV can only access its own observation but can not get the states of other UAVs. 
The observation vector of each UAV consists of three parts: $\bm{o}^{t} = [\bm{o}^{t}_{z}, \bm{o}^{t}_{g}, \bm{o}^{t}_{v}]$, where $\bm{o}^{t}_{z}$ denotes the images of the surrounding environment obtained by the vision sensor, $\bm{o}^{t}_{g}$ denotes its relative goal position, and $\bm{o}^{t}_{v}$ denotes its current velocity.
Given the partial observation $\bm{o}^{t}$, each UAV independently and simultaneously computes an action $\bm{a}^{t}$ which sampled from the policy $\pi$ shared by all UAVs:
$$
    \bm{a}^{t} \thicksim \pi_{\theta}(\cdot | \bm{o}^{t}) \eqno{(1)}
$$
where $\theta$ denotes the policy's parameter vector. The action $\bm{a}^{t}$ represents the velocity command of the UAV. 
It will drive the UAV to approach its goal and avoid collisions within the time horizon $\Delta t$ until it receives the next observation $\bm{o}^{t+1}$.

\subsection{Reinforcement Learning Setting}

As mentioned in Section III.A, the multi-UAV collision avoidance problem is a POMDP, described as a 6-tuple $(\mathcal{S}, \mathcal{A}, \mathcal{T}, \mathcal{R}, \Omega, \mathcal{O})$.
In addition, since the path planning of each UAV is completely decentralized, the multi-UAV state transition model $\mathcal{T}$ determined by UAVs' kinematics and dynamics is not needed.
In our formulation, each UAV makes decisions only through its own observation, and improves the policy through the reward function.
In the following, we give the details of the observation space, the action space and the reward function.

\subsubsection{Observation space}
The $\bm{o}^{t}_{z}$ refers to depth images taken from the camera placed in front of the UAVs at timestep $t$, which contains information relating to the distance of the surfaces of scene objects from a viewpoint. We stack three depth frames together to infer visual and temporal features further.
The $\bm{o}^{t}_{g}$ refers to the goal position in the body coordinate system because we use body frame commands (forward $\bm{v}_{x}^{cmd}$, climb $\bm{v}_{z}^{cmd}$ and steering $\bm{v}_{\omega}^{cmd}$) to control each UAV.
The $\bm{o}^{t}_{v}$ refers to the velocity of the UAVs at the current time: $\bm{o}^{t}_{v} = [\bm{v}^{t}_{x}, \bm{v}^{t}_{z}, \bm{v}^{t}_{\omega}]$.

\subsubsection{Action space}
To make the movement of the UAV more smooth and controllable, the action space is a set of limited velocities in continuous space.
Action $a=[\bm{v}_{x}^{cmd}$, $\bm{v}_{z}^{cmd}$, $\bm{v}_{\omega}^{cmd}]$ calculated from the policy network $\pi(s)$ consists of two linear velocities and an angular velocity.
In this work, we limit the range of the forward velocity $\bm{v}_{x}^{cmd} \in (0.0, 2.0)$ m/s, the climb velocity $\bm{v}_{z}^{cmd} \in (-0.5, 0.5)$ m/s and the steering velocity $\bm{v}_{\omega}^{cmd} \in (-0.5, 0.5)$ rad/s. Note that backward velocity(i.e., $\bm{v}_{x}^{cmd} < 0$ m/s) is not allowed. This is because each UAV is only equipped with a camera in the front, which can not observe behind and can not help avoid collisions if the UAV moves backwards.

\subsubsection{Reward function}
Our objective is to maximize the path efficiency and safety of all UAVs. A reward function is designed to facilitate the UAVs to learn a policy to achieve this objective:
$$
    \bm{r}^{t}=\bm{r}^{t}_{goal}+\bm{r}^{t}_{avoid} \eqno{(2)}
$$

The reward $\bm{r}^{t}_{goal}$ is designed to help each UAV to reach its goal. If the UAV is within 0.5 meters from the goal at the current timestep, it will receive a constant reward $\bm{r}_{arrival}$, otherwise $\bm{r}^{t}_{goal}$ is the change of the distance to the goal from the previous timestep to the current timestep, scaled by a rewarding  weight $\bm{\omega}_{goal}$, a hyperparameter to encourage efficient paths:
$$
    \bm{r}^{t}_{goal}=\left\{
        \begin{array}{l}
            \bm{r}_{arrival} \quad \quad \quad \quad \quad \quad \quad \quad if \  \Vert \bm{p}_{i}^{t}-\bm{g}^{t}_{i} \Vert < 0.5 \\
            \bm{\omega}_{goal} \cdot (\Vert \bm{p}_{i}^{t-1}-\bm{g_{i}}^{t} \Vert - \Vert \bm{p}_{i}^{t}-\bm{g}_{i}^{t} \Vert)  \ otherwise
        \end{array}
        \right.
    \eqno{(3)}
$$

The reward $\bm{r}^{t}_{avoid}$ is designed to encourage each UAV to avoid collision. We assume that the collision radius of each UAV is $R$. The UAV is penalized by $\bm{r}_{collision}$ when it collides with other UAVs in the environment. Otherwise, the UAV is punished when the minimum distance in the depth image $\bm{d}^{t}_{min}$ is less than the safe distance $\bm{d}_{safe}$, scaled by a penalty weight $\bm{\omega}_{avoid}$:
$$
    \bm{r}^{t}_{avoid}=\left\{
        \begin{array}{l}
            \bm{r}_{collision} \quad \quad \quad \quad \quad \quad \quad if \ \Vert \bm{p}^{t}_{i}-\bm{p}^{t}_{j} \Vert < 2R \\
            \bm{\omega}_{avoid} \cdot {max}(\bm{d}_{safe}-\bm{d}^{t}_{min}, 0) \quad otherwise
        \end{array}
        \right.
        \eqno{(4)}
$$
We set $\bm{r}_{arrival}=50$, $\bm{r}_{collision}=-10$, $\bm{\omega}_{goal}=3$, $\bm{\omega}_{avoid}=-0.05$ and $\bm{d}_{safe}=5$ during the training procedure. 

\begin{figure}
    \centering
    \includegraphics[width=240pt]{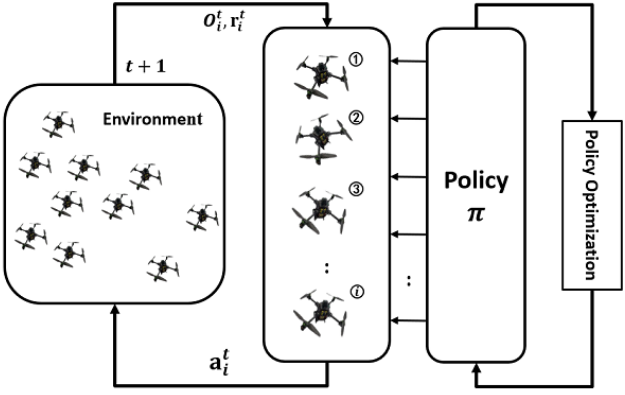}
    \caption{An overview of our approach. In an environment with $N$ UAVs, at each time step, each UAV receives it observation $\bm{o}_{i}^{t}$ and reward $\bm{r}_{i}^{t}$ from the environment, and executes an action $\bm{a}_{i}^{t}$ following the policy $\pi$. The policy $\pi$ is shared cross all UAVs and updated by a DRL algorithm.}
    \label{fig3}
\end{figure}

\subsection{Network Architecture}
Since it is challenging to train agents to solve the control task directly from the pixels of high-dimensional images via model-free reinforcement learning, we design a neural network based on $\textbf{SAC+AE}$ [28], a robust and efficient method of learning policy from high-dimensional images, for training our multi-UAV navigation policy $\pi_{\theta}$. Its architecture is shown in Fig. 2. A deterministic autoencoder is used to extract visual and temporal features from three consecutive depth images $\bm{o}_{z}^{t}$. The encoder consists of 4 convolutional layers and a fully connected layer that maps the input depth images to a 50-dimensional latent representation. The first convolutional layer consists of 32 two-dimensional filters with kernel size = 3, stride = 2 over the input depth images, and ReLU nonlinearities are applied. The rest convolutional layers convolve 32 two-dimensional filters with kernel size = 3, stride = 1, again followed by ReLU nonlinearities. The fully connected (FC) layer outputs a 50-dimensional latent representation with layer normalization. The decoder consists of an FC layer and four deconvolutional layers with the same setup as the encoder, which reconstruct the latent representation back to the original image. The latent representation is concatenated with the other two observations ($\bm{o}_{g}^{t}$ and $\bm{o}_{v}^{t}$), and as input for actor and critic. The actor consists of a 3-layer multi-layer perceptron (MLP) with a 1024-dimensional hidden layer and outputs 3-dimensional (corresponding to three actions) mean vector and covariance matrix of parameterized gaussian action distributions. The output of the critic (Q-function) is represented by a 3-layer MLP with 1024 units. To improve the policy's performance, we prevent the actor’s gradients from updating the convolutional encoder.

\subsection{Multi-UAV Policy Training}
\subsubsection{Training method}
To learn the effective collision-free navigation policy of multi-UAV, we deploy an extended $\textbf{SAC+AE}$ [28] in our multi-UAV system. As shown in Fig. 3, our method adopts the paradigm of centralized learning and distributed execution [29]. During training, the policy is learned from the trajectories collected by all UAVs. At each time step $t$, the $i$th UAV accesses observation $\bm{o}_i^t$, and executes action $\bm{a}_i^t$ generated from the shared policy $\pi_\theta$. As shown in Algorithm 1, the training process mainly consists of two stages. One is that multiple UAVs execute the policy simultaneously and store their trajectories $(\bm{o}_{i}^t,\bm{a}_{i}^t,r_{i}^t,\bm{o}_{i}^{t'},d_{i}^{t})$ in a replay buffer $\mathcal{B}$. The other is to update the policy and autoencoder by sampling a batch of trajectories $(\bm{o},\bm{a},r,\bm{o}^{'},d)$ from $\mathcal{B}$.

\begin{algorithm}[h]
    Initialize actor, critic, encoder, decoder and replay buffer $\mathcal{B}$ \;
    \While{Episode $<$ max episodes}{
      \For{timestep t = 1, 2, ... }{
        \For{UAV i = 1, 2, ..., $N$}{
          \While{t $<$ max time steps $T_{max}$ or terminal is not done}{
            Select action $\bm{a}_{i}^{t} \thicksim \pi_{\theta}(\cdot | \bm{o}_{i}^{t})$\;
            Receive reward $r_{i}^t$, observation $\bm{o}_{i}^{t'}$ and done signal $d_{i}^{t}$\;
            Store $(\bm{o}_{i}^t,\bm{a}_{i}^t,r_{i}^t,\bm{o}_{i}^{t'},d_{i}^{t})$  in $\mathcal{B}$ \;
          }}
          }
          \While{cnt $<$ update times}{
          Randomly sample batch of trajectories $(\bm{o},\bm{a},r,\bm{o}^{'},d)$ from $\mathcal{B}$ \;
          Update actor, critic, encoder by $J(Q)$ and $J(\pi)$ \;
          Update encoder and decoder by $J(RAE)$ \;
          }
      }
    \caption{Our method for learning the collision-advoidance policy}
  \end{algorithm}

The objective functions of the updated policy are as follows. In the policy evaluation step, the objective is to fit Q-function [30]:
$$
    J(Q)=\mathbb{E}_{(\bm{o},\bm{a},r,\bm{o}^{'})\thicksim\mathcal{B}}\left[(Q(\bm{o},\bm{a})-r-\gamma\bar{V}(\bm{o}^{'}))^{2}\right] 
    \eqno{(5)}
$$
$$
    \bar{V}(\bm{o})=\mathbb{E}_{\bm{a} \thicksim \pi_{\theta}}\left[(\bar{Q}(\bm{o},\bm{a})-r-\alpha \log \pi(\bm{a}|\bm{o})\right] 
    \eqno{(6)}
$$
where $\bar{Q}$ is the target Q-function parametrized by a weight vector obtained from an exponentially moving average of the Q-function weights to stabilize training.

In the policy improvement step, the objective is to minimize the KL divergence between the Boltzmann distribution induced by the Q-function and policy:
$$
    J(\pi)=\mathbb{E}_{\bm{o}\thicksim\mathcal{B}}\left[D_{KL}(\pi(\cdot|\bm{o})\| \mathcal{Q}(\bm{o},\cdot))\right] 
    \eqno{(7)}
$$
where $\mathcal{Q}(\bm{o},\cdot)\propto$ exp $\{\frac{1}{\alpha} Q(\bm{o},\cdot)\}$

We adopt a deterministic autoencoder RAE (Regularization Autoencoder) [31] to update the encoder $p_{\phi}$ and decoder $g_{\varphi}$ with the following objective:
$$
    J(RAE)=\mathbb{E}_{\bm{x}}\left[\log p_{\phi}(\bm{x}|\bm{z})+\lambda_{\bm{z}}\|\bm{z}\|^{2}+\lambda_{\phi}\|\phi\|^{2}\right] 
    \eqno{(8)}
$$
where $\bm{x}$ is the image extracted from the observation $\bm{o}$, $\bm{z}=g_{\varphi}(\bm{x})$, and $\lambda_{\bm{z}}$, $\lambda_{\phi}$ are hyperparameters. Here we set $\lambda_{\bm{z}} = 10^{-6}$ and $\lambda_{\phi} = 10^{-7}$.

\begin{figure}
    \centering
    \includegraphics[width=240pt]{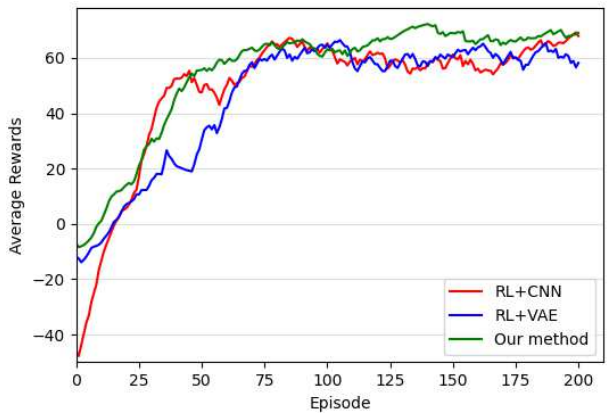}
    \caption{Comparison of the average rewards during training (red line for RL+CNN, blue line for RL+VAE and green line for our method).}
    \label{fig4}
\end{figure}

\section{EXPERIMENTS AND RESULTS}
\subsection{Simulation Environment and Experiments Setup}
To achieve robust multi-UAV collision avoidance policy learning, we implemented our algorithm with ROS noetic and built a scenario with multi-UAV on the Airsim simulator [32] (as shown in Fig. 1). We built our policy model based on Pytorch and trained it on a PC with Intel i9-11900k and NVIDIA Quadro RTX 4000. 

In each episode, the proper initial position and goal point of each UAV are randomly generated in three-dimensional space to allow the UAV to fully explore its high-dimensional observation space and potentially improve the efficiency and robustness of the learning policy. The hyper-parameters for policy training are listed in Table I.

\begin{table}[h]
    \caption{Hyperparameters for policy trainning.}
    \label{APPENDIX_1}
    \begin{center}
        \begin{tabular}{p{5cm} p{2cm}}
            \toprule
            Parameters name& Value \\
            \midrule
            \multirow{1}[1]{*}{Replay buffer $\mathcal{B}$ capacity} & 20000 \\
            \multirow{1}[1]{*}{Batch size} & 128 \\
            \multirow{1}[1]{*}{Max episodes} & 200 \\
            \multirow{1}[1]{*}{Update times} & 400 \\
            \multirow{1}[1]{*}{Discount $\gamma$} & 0.99 \\
            \multirow{1}[1]{*}{Optimizer} & Adam \\
            \multirow{1}[1]{*}{Critic learning rate} & $10^{-3}$ \\
            \multirow{1}[1]{*}{Critic target update frequency} & 2 \\
            \multirow{1}[1]{*}{Critic Q-function soft-update rate $\tau_{Q}$} & 0.01 \\
            \multirow{1}[1]{*}{Critic encoder soft-update rate $\tau_{enc}$} & 0.05 \\
            \multirow{1}[1]{*}{Actor learning rate} & $10^{-3}$ \\
            \multirow{1}[1]{*}{Actor update frequency} & 2 \\
            \multirow{1}[1]{*}{Actor log stddev bounds} & [-10,2] \\
            \multirow{1}[1]{*}{Autoencoder learning rate} & $10^{-3}$ \\
            \bottomrule
            \end{tabular}                                                                                          
    \end{center}
 \end{table}

\begin{figure}
    \centering
    \includegraphics[width=235pt]{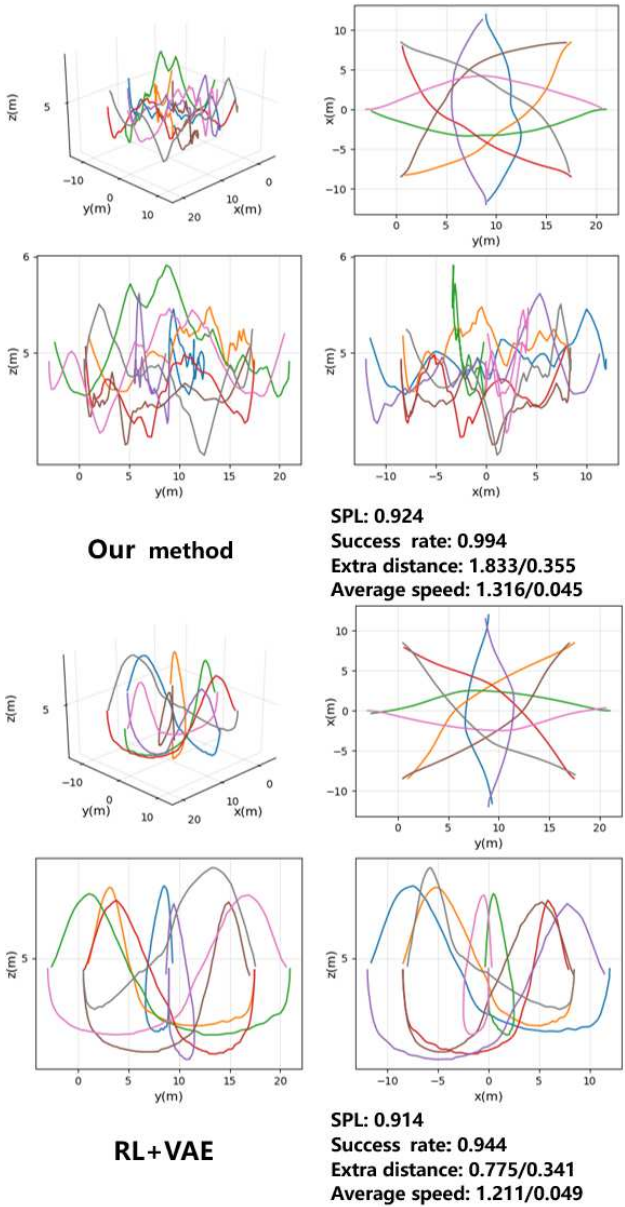}
    \caption{Illustration of the trajectories of our method and RL+VAE under circle scenario (radius = 12 m, number of UAVs = 8) in perspective drawing and three-view drawing. We use different colors to represent trajectories of different UAVs. The performance metrics are shown as mean/std among all test cases.}
    \label{fig5}
\end{figure}

\subsection{Performance Metrics and Experiment Scenarios}
To compare the performance of our method with other methods, we provide the following quantitative evaluation metrics:
\begin{itemize}
\item Success Rate:
The percentage of UAVs that successfully reach their own goals in a limited time without any collisions. 
\item SPL(Success weighted by Path Length):
An extended evaluation of the agent’s navigation performance [33] for multi-UAV across the test scenario in $N$ UAVs and $M$ episodes as follows: 
$$
    \frac{1}{N} \frac{1}{M} \sum_{i = 1}^{N}\sum_{j = 1}^{M}S_{i,j}\frac{l_{i,j}}{max(p_{i,j}, l_{i,j})}   
    \eqno{(9)}
$$
where $l_{i,j}$ is the shortest-path distance from the $i$th UAV’s initial position to the goal position in episode $j$, $p_{i,j}$ is the length of the path actually taken by the $i$th UAV in this episode and $S_{i,j}$ is a binary indicator of success or failure in this episode.
\item Extra Distance:
The average extra distance of UAVs travels compared to the straight-line distance between the initial and goal positions.
\item Average Speed:
The average speed of all UAVs during testing. 
\end{itemize}

We employ two types of testing scenarios in our experiment: 
\begin{itemize}

\item Random Scenarios:
The initial position and goal position of each UAV are randomly generated in a reasonable area. According to the size of the area, we divide the scene into three types, namely large scene (around 0.04 UAV/$m^3$), medium scene (around 0.06 UAV/$m^3$) and small scene (around 0.1 UAV/$m^3$).

\item Cycle Scenarios:
All UAVs are located uniformly on a circle at a specific altitude. They need to reach their goal set on the opposite side of the circle.

\end{itemize}

Note that our policy is trained on medium size of random scenarios. Other test scenarios are never seen during the training process.

\begin{table*}
    \caption{Performance metrics (as mean/std) evaluated for different methods on the scenarios with varied scene sizes.}
    \label{table_1}
    \begin{center}
        \begin{tabular}{cccccc}
            \toprule
            Scene size& Method & Success Rate & SPL & Extra distance & Average speed\\
            \midrule
            \multirow{3}[1]{*}{Small} & Our method & \textbf{0.895}  & 0.715 & 2.035/2.270 & 0.914/0.130 \\
                  & RL+CNN & 0.764 & 0.637 & 1.505/1.154 & 0.715/0.070 \\
                  & RL+VAE & 0.818 & \textbf{0.745} & \textbf{0.686/0.494} & \textbf{0.967/0.099} \\
            \multirow{3}[1]{*}{Medium} & Our method & \textbf{0.934}  & \textbf{0.791} & 1.784/1.774 & 0.991/0.104\\
                  & RL+CNN & 0.829 & 0.717 & 1.499/1.080 & 0.771/0.065 \\
                  & RL+VAE & 0.851 & 0.785 & \textbf{0.777/0.607} & \textbf{1.038/0.090} \\
            \multirow{3}[1]{*}{Large} & Our method & \textbf{0.952}  & \textbf{0.852} & 1.334/1.202 & \textbf{1.066/0.094} \\
                  & RL+CNN & 0.886 & 0.774 & 1.663/1.133 & 0.773/0.072 \\
                  & RL+VAE & 0.900 & 0.833 & \textbf{0.886/0.536} & 1.063/0.082\\
            \bottomrule
            \end{tabular}                                                                                                                                  
    \end{center}
 \end{table*}

 \begin{figure}[h]
    \centering
    \includegraphics[width=157pt]{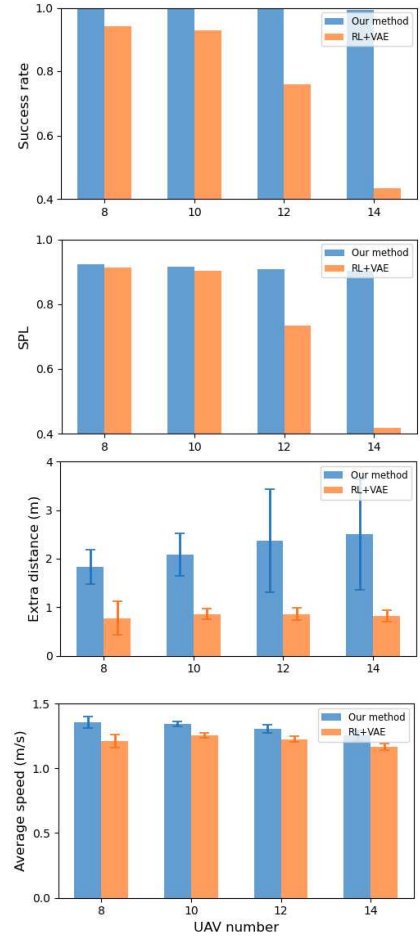}
    \caption{Illustration of the performance metrics (Success rate, SPL, Extra distance and Average speed) of our method and RL+VAE under circle scenario (radius = 12 m, number of UAVs = 8, 10, 12, 14).}
    \label{fig6}
\end{figure}

 \begin{figure*}
    \centering
    \includegraphics[width=505pt]{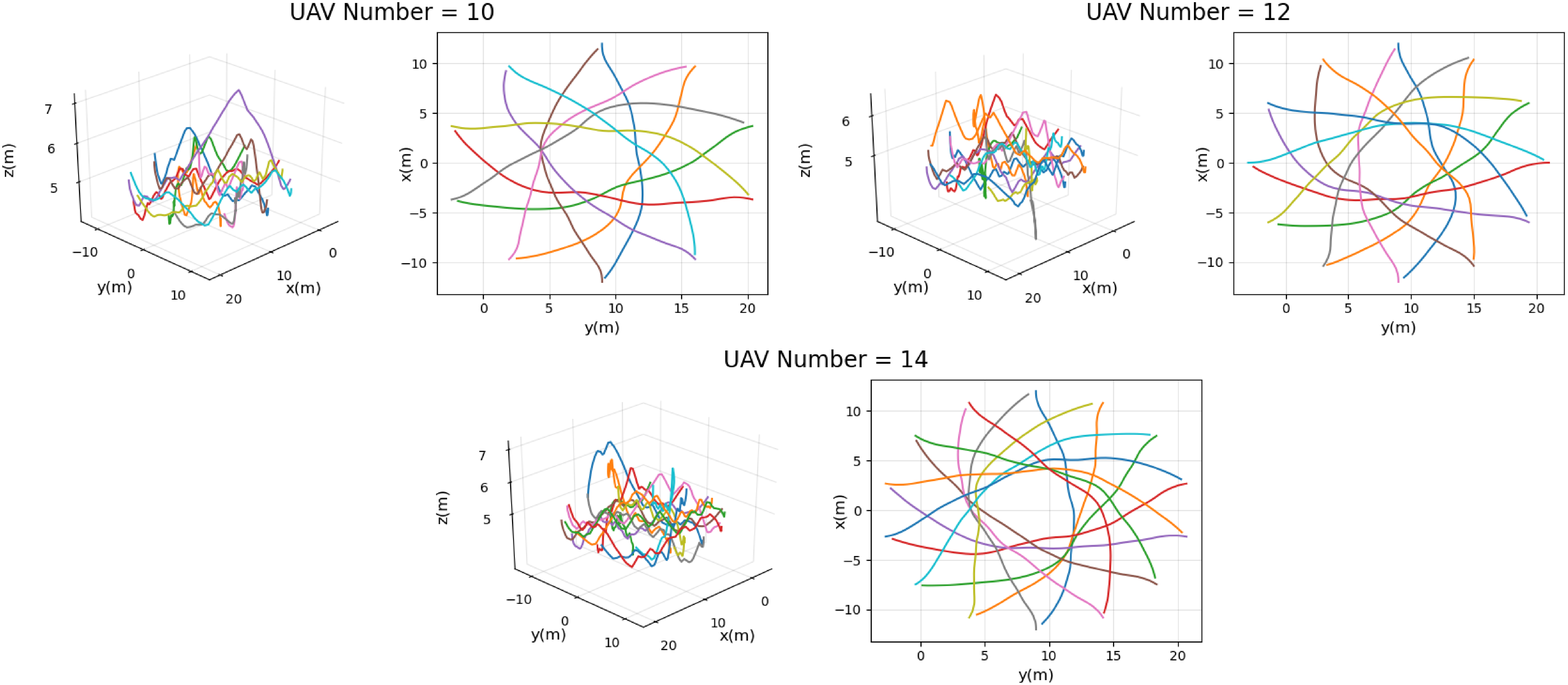}
    \caption{Illustration of the trajectories of our method under circle scenario (radius = 12 m, number of UAVs = 10, 12, 14) in perspective drawing and birds-eye-view drawing. We use different colors to represent trajectories of different UAVs.}
    \label{fig7}
\end{figure*}

\subsection{Experiment Results on Various Scenarios}
We compare our policy with different DRL-based methods. One of them combines a DRL algorithm and convolutional neural network (RL+CNN) [34], and the other uses Variational Autoencoder (VAE) to compress image information combined with the policy-based DRL model (RL+VAE) [35]. For RL+VAE, we use our proposed policy to collect images with multi-UAV to train the variational autoencoder.

We separately trained different methods on the medium size of random scenarios and obtained the learning curves shown in Fig. 4. We can see that all these methods converge during training, and our proposed method converge to higher reward.

Tables II shows the performance evaluated for different methods on different sizes of random scenarios after 1000 episodes, and the best results are highlighted in \textbf{bold}. It can be seen that our method shows excellent consistency in performance, i.e., with the increase of scene size, the success rate, SPL and average speed are gradually increasing, and the extra distance is gradually decreasing.  In the three different scenarios, although our method adds some extra distance for better collision avoidance, it achieves a higher success rate than other methods. In the simple scene (large scene), our method is 6.6$\%$ and 5.2$\%$ higher than RL+CNN and RL+VAE respectively, while in the complex scene (small scene), our method is more competitive, 13.1$\%$ and 7.7$\%$ higher than RL+CNN and RL+VAE respectively. Notable is,  compared with other methods, our method obtains the highest success rate and the maximum with the fastest average speed in the large scene, and achieves the best success rate (approximately 90$\%$) in the small scene, which indicate that our method is more suitable for practical group tasks. Furthermore, our method and RL+VAE perform better than RL+CNN, which demonstrates that auto-encoder is helpful for policy learning.

We also test the performance evaluated for different methods on circle scenario (radius = 12 m, number of UAVs = 8) and find that RL+CNN behaves terribly during testing, therefore it is not added to the subsequent comparative experiments. We visualize the trajectories of our method and RL+VAE under the circle scenario and show the performance metrics, as shown in Fig. 5.
Our method demonstrates better navigation performance, except for the metric of extra distance. This is because each UAV adjusts its planning path more frequently to avoid collision compared to RL+VAE. We will improve our method to shorten extra distance while maintaining the success rate in future work.

Scalability and robustness are essential in multi-UAV systems. To test the performance of scalability and robustness, we increase the number of UAVs in the circle scenario and make a quantitative analysis. As shown in Fig. 6, with the number of the UAVs increases (the radius of the circle scenario is fixed at 12 m), although the performance of our method decreases slightly, the success rate and SPL remain at a high level (the success rate is close to 100$\%$ and the SPL is higher than 90$\%$). The extra distance and average speed are increasing and decreasing respectively to avoid UAVs collisions in the scene. The extra distance and average speed change little with RL+VAE, but the success rate and SPL decrease seriously with the increase of the number of UAVs, especially when there are more than 12 UAVs in the scene. These results show that our method has better robustness and scalability than RL+VAE, therefore our method can be deployed to large-scale multi-UAV systems.

Moreover,  through observation, we find that in a circle scenario (Fig. 5 and Fig. 7), the trajectories of UAVs form relatively regular shapes from a bird's-eye view.  It demonstrates that this completely communication-free vision-based decentralized collision-avoidance policy enables multiple UAVs certain swarm cooperative behavior emerges in the system.

\section{CONCLUSIONS}
We present an effective, vision-based multi-UAV collision-avoidance policy learning method. We show that a vision-based model is enough to achieve multi-UAVs collision avoidance without spatial representation. In addition, the experimental results show that our proposed method can offer better robust and scalable performance in different scenarios compared to prior approaches.

Though our proposed method is forward-looking, there are many ways to improve the performance. Current training scenarios do not include any static or dynamic obstacles, we will add them in future work. We consider deploying multiple cameras on each UAV to enhance its environment perception [36]. The most important thing for future work is extending to real-world scenes, where we need to use other techniques to guarantee robustness and safe. It would be useful to combine our methods with model predictive control (MPC) [37], a powerful model-based approach for solving complex quadrotor control problems.

\addtolength{\textheight}{-12cm}   




\end{document}